\documentclass[11pt,a4paper]{article}
\usepackage[hyperref]{acl2021}
\usepackage{times}
\usepackage{latexsym}

\usepackage{microtype}

\usepackage{soul}
\usepackage{tablefootnote}
\usepackage[normalem]{ulem}
\usepackage{multirow}
\usepackage{makecell}
\usepackage{soul}
\usepackage{enumitem}
\usepackage{adjustbox}
\usepackage{amsmath}
\usepackage{amssymb}
\usepackage{array}
\usepackage{graphicx}
\usepackage{comment}
\usepackage{booktabs}
\usepackage{xcolor,colortbl} 
\usepackage{algorithm}
\usepackage{algorithmic}

\usepackage{bbm}
\usepackage{tabularx}
\usepackage{hyperref}

\usepackage{graphicx}
\usepackage{subcaption}
\usepackage{arydshln}
\usepackage{array}
\usepackage{booktabs}
\usepackage{pifont}

\aclfinalcopy 
\def\aclpaperid{513} 

\title{Answering Ambiguous Questions through Generative Evidence Fusion and Round-Trip Prediction}

\author{
 \makecell{Yifan Gao$^\dag$\Thanks{~Work done during an internship at AWS AI.}, Henghui Zhu$^\ddag$, Patrick Ng$^\ddag$, Cicero Nogueira dos Santos$^\ddag$, Zhiguo Wang$^\ddag$, \\ Feng Nan$^\ddag$, Dejiao Zhang$^\ddag$, Ramesh Nallapati$^\ddag$, Andrew O. Arnold$^\ddag$, Bing Xiang$^\ddag$} \\
 {$^\ddag$AWS AI}  \quad  {$^\dag$ The Chinese University of Hong Kong} \\
 \small
 \tt{$^\dag$yifangao95@gmail.com \quad $^\ddag$\{henghui,patricng,cicnog,zhiguow\}@amazon.com} \\
 \small{\tt $^\ddag$\{nanfen,dejiaoz,rnallapa,anarnld,bxiang\}@amazon.com} \\
}

\date{}

\begin{document}
\newcommand{\tocite}[1]{{[\hl{CITE: #1]}}}
\newcommand{\todo}[1]{{[\hl{TODO: #1}]}}
\newcommand{\yifan}[1]{{[\hl{Yifan: #1}]}}

\newcommand{\modelnameshort}{\textsc{Refuel}}
\newcommand{\modelname}{\textsc{Refuel}}
\newcommand{\sota}{{state-of-the-art}}
\newcommand{\ambignq}{\textsc{AmbigQA}}
\newcommand{\ambigqa}{\textsc{AmbigQA}}
\newcommand{\nqopen}{\textsc{NQ-open}}
\newcommand{\triviaqa}{{TriviaQA}}
\newcommand{\spanseqgen}{\textsc{SpanSeqGen}}

\newcommand{\Fanswer}{$\mathrm{F1}_\text{ans}$}
\newcommand{\Fbleu}{$\mathrm{F1}_\text{BLEU}$}
\newcommand{\Fedit}{$\mathrm{F1}_\text{EDIT-F1}$}

\newcommand{\Prompt}{Prompt question}
\newcommand{\prompt}{prompt question}
\newcommand{\prompts}{prompt questions}
\newcommand{\dev}{development}
\newcommand{\devshort}{development}

\newcommand{\QDF}{\textsc{QdF}}
\newcommand{\QGP}{\textsc{QgP}}
\newcommand{\TDP}{\textsc{TdP}}
\maketitle

\begin{abstract}

In open-domain question answering, questions are highly likely to be ambiguous because users may not know the scope of relevant topics when formulating them. 
Therefore, a system needs to find possible interpretations of the question, and predict one or multiple plausible answers.
When multiple plausible answers are found, the system should rewrite the question for each answer to resolve the ambiguity.
In this paper, we present a model that aggregates and combines evidence from multiple passages to adaptively predict a single answer or a set of question-answer pairs for ambiguous questions.
In addition, we propose a novel round-trip prediction approach to iteratively generate additional interpretations that our model fails to find in the first pass, and then verify and filter out the incorrect question-answer pairs to arrive at the final disambiguated output.
Our model, named \modelnameshort, achieves a new state-of-the-art performance on the \ambignq\ dataset, and shows competitive performance on \nqopen\ and \triviaqa.
The proposed round-trip prediction is a model-agnostic general approach for answering ambiguous open-domain questions, which improves our \modelnameshort\ as well as several baseline models.
We release source code for our models and experiments at \url{https://github.com/amzn/refuel-open-domain-qa}.
\end{abstract}

\section{Introduction}
\begin{figure}[t!]
\footnotesize
\begin{tabular}{p{0.95\columnwidth}}
\hline\hline
\textbf{Prompt Question} ({Google search query}): What's the most points scored in an NBA game? \\
\textbf{Disambiguated QA Pairs:} \\
$\mathbf{Q}_1$: What's the most points scored in an NBA game by combined team? / $\mathbf{A}_1$: 370 \\
$\mathbf{Q}_2$: What's the most points scored in an NBA game by a single team? /
$\mathbf{A}_2$: 186 \\
$\mathbf{Q}_3$: What's the most points scored in an NBA game by an individual? /
$\mathbf{A}_3$: 100 \\
\hline
\textbf{Relevant Wikipedia Page 1}:
The highest-scoring regular season game is the triple-overtime game between ... the two teams combined to score \hl{370} points, with the pistons defeating the nuggets \hl{186}–184 ... \\
\textbf{Relevant Wikipedia Page 2}: 
Wilt Chamberlain scored an nba-record \hl{100} points ... \\
\hline\hline                
\end{tabular}
\caption{An example from the \ambignq\ \cite{min-etal-2020-ambigqa} dataset. The \textbf{Prompt Question} is gathered from Google search queries and has three interpretations upon reading Wikipedia. \textbf{Disambiguated QA Pairs} are the full set of acceptable answers, paired with the disambiguated rewriting of the prompt question. 
}
\label{fig:example}
\end{figure}

Open-domain Question Answering (QA) is the task of answering questions using a collection of passages with diverse topics \cite{chen-etal-2017-reading, Guu2020REALMRL, karpukhin2020dense}.
Open-domain questions are highly likely to be ambiguous because people may not have the knowledge of relevant topics when formulating them. 
For example, in Figure \ref{fig:example}, the prompt question \emph{``What's the most points scored in an NBA game?''} is ambiguous because the \textit{score} in this question could be interpreted as \emph{the combined score in a game} ($\textrm{Q}_1\textrm{A}_1$), \emph{score from a single team} ($\textrm{Q}_2\textrm{A}_2$), or \emph{score from an individual player} ($\textrm{Q}_3\textrm{A}_3$).
Therefore, a system needs to adaptively predict a single answer, or a set of equally plausible answers when the question has multiple interpretations.
When a set of multiple answers is predicted, an unambiguous rewriting of the question that leads to each answer should also be provided to clarify each interpretation.

\newcite{min-etal-2020-ambigqa} decompose this problem into two subtasks.
Given the prompt question and Wikipedia passages, the first subtask, \textbf{Answer Prediction}, consists in predicting one or several plausible answers, depending on whether this question is ambiguous or not. 
If multiple answers are predicted, the second subtask, \textbf{Question Disambiguation}, requires generating a disambiguated question for each of the plausible answers.
They propose \spanseqgen, which first retrieves and reranks passages using the prompt question, and then adopts a BART pre-trained sequence-to-sequence model \cite{lewis-etal-2020-bart} to generate all plausible answers, conditioned on the concatenation of the prompt question and top 8 passages.
For the question disambiguation subtask, based on BART, they first pre-train a question generation model on \nqopen\ \cite{kwiatkowski2019natural}, a large-scale open-domain QA dataset, to generate the question given the answer and top 8 passages. Then they fine-tune it as a question disambiguation model to generate the disambiguated question conditioned on the prompt question, answer, and passages.

There are three main drawbacks to \spanseqgen. 
Firstly, a complete coverage of all relevant passages is essential for predicting all plausible answers of the ambiguous question. 
However, \spanseqgen\ only takes 8 passages for answer prediction so some of the most informative passages might be excluded.
Secondly, for the question disambiguation subtask, there is a mismatch between question generation pre-training on \nqopen\ and question disambiguation fine-tuning on \ambignq\ -- there is no question to disambiguate in question generation pre-training, which makes the pre-training task somewhat misaligned with fine-tuning.
Thirdly, \spanseqgen\ predicts a much smaller average number of answers compared to the ground truth data (1.17 \textit{vs.} 2.19).

To address these issues, we propose \modelnameshort, Round-trip Evidence FUsion via gEneration with retrievaL, a new framework for answering ambiguous open-domain questions.
To ensure a broad coverage of relevant knowledge of the question, \modelnameshort\ reads 12 times more passages (100 in our experiments) than \spanseqgen\ by 
using Fusion-in-Decoder \cite{Izacard2020LeveragingPR} that processes each passage individually in the encoder, and then fused their encodings together in the decoder.
For the question disambiguation subtask, we propose a \textit{token-deletion pre-training task} to transform \nqopen\ into an ``ambiguous'' QA setting by randomly deleting an informative span for each question.
Thus, pre-training and fine-tuning tasks are well aligned.
Additionally, we add an \textit{insertion-based weighted loss} to emphasize the newly inserted tokens in the disambiguated question, which helps the model on learning to resolve the ambiguity.
Finally, we propose a \textit{round-trip prediction approach} to find additional interpretations that \modelnameshort\ fails to predict in the first pass.
We continuously feed the generated questions into \modelnameshort\ until there are no new answers predicted from our model.
While this round-trip prediction can improve the recall of answers, we refine the quality of predicted QA pairs by filtering them with the conditional probability of the answers estimated by an answer-generation model.

Our \modelnameshort\ achieves a new state-of-the-art on the \ambignq\ dataset, outperforming the previous best model \spanseqgen\ by 9.1\% in answer prediction F1 and 4.4\% in Edit-F1 score for question disambiguation.
When directly doing inference on \nqopen\ and \triviaqa, \modelnameshort\ not only predicts the single answer precisely but also finds multiple interpretations if the question is ambiguous.
Moreover, human evaluation shows that \modelnameshort\ can correctly generate more QA pairs on all three datasets.
Finally, the proposed round-trip prediction is a model-agnostic general approach for answering ambiguous questions, which improves our \modelnameshort\ as well as several baseline models up to 3.7\% for the overall performance.

\begin{figure*}[t!]
\centering
\includegraphics[width=\textwidth]{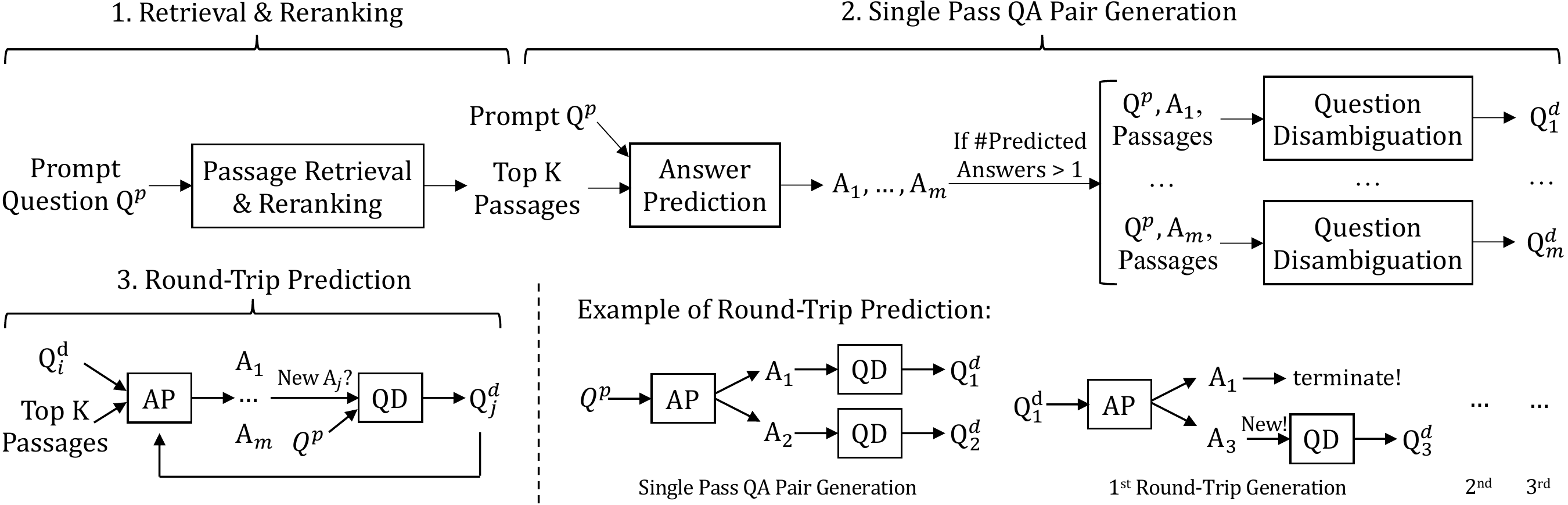}
\caption{Overall Pipeline of \modelnameshort. 
\modelnameshort\ firstly retrieves question-relevant passages (Section \ref{sec:rrrk}). 
Then it generates first-pass QA pairs through the Answer Prediction (AP) module and Question Disambiguation (QP) module (Section \ref{sec:single-detail}).
Finally, generated disambiguated questions $Q^d$ are further taken as the input of our pipeline to find more interpretations (Round-Trip Prediction). If the generated question $Q^d$ still has multiple interpretations, the newly predicted answers will receive their own questions (Section \ref{sec:round}). 
}
\label{fig:model}
\end{figure*}

The main contributions of this work, which are fundamental to significantly push the \sota\ in answering ambiguous questions, can be summarized as follows:
\vspace{-0.08in}
\begin{enumerate}[noitemsep]
    \item We present an evidence aggregation approach that can effectively use a large number of passages to uncover more candidate interpretations of the ambiguous question.
    \item We propose a token-deletion pre-training task to reduce the mismatch between pre-training and fine-tuning for question disambiguation. The insertion-based weighted loss further helps to capture answer-relevant constraints.
    \item We propose a round-trip prediction approach to find more interpretations missed in the first prediction pass, which we further refine using a conditional-probability-based filtering approach.
\end{enumerate}

\section{\modelnameshort}
\modelnameshort\ answers questions through a three-step process illustrated in Figure \ref{fig:model}:
\begin{enumerate}[noitemsep]
    \item The \textit{Passage Retrieval \& Reranking} module retrieves question-relevant passages from the whole Wikipedia corpus. Then the retrieved passages are further reranked (Sec. \ref{sec:rrrk}). 
    \item Taking the reranked passages and the prompt question as input, our single pass QA pair generation model makes the first prediction pass to predict a single answer or a set of disambiguated QA pairs (Sec. \ref{sec:single-general}). 
    \item Our proposed \textit{Round-Trip Prediction} can find more interpretations missed in the first prediction pass, which we further refine using a conditional-probability-based filtering approach (Sec. \ref{sec:round}). 
\end{enumerate}

\subsection{Passage Retrieval \& Reranking}\label{sec:rrrk}
We use Dense Passage Retriever (DPR) \cite{karpukhin2020dense} for retrieval.
First, we split all Wikipedia pages into 100-token passages, resulting in 24M passages in total.
Then DPR maps all passages into $d$-dimensional vectors, computes the representation of the prompt question, and retrieves N passages whose vectors are closest to the question vector (we use N=1000).

After retrieving N passages for the prompt question, we fine-tune BERT \cite{devlin-etal-2019-bert} to rerank these passages. 
Taking the concatenation of the prompt question and each passage as input, the reranker allows a token-level cross-attention between the prompt question and passages.
The relevance score is then derived by taking the \texttt{[CLS]} vector of the input sequence into a linear layer.
After reranking, the QA pair generation model takes the top K passages as inputs (we use K=100).

\subsection{Single Pass QA Pair Generation}\label{sec:single-general}

The single pass QA pair generation step includes an Answer Prediction module and a Question Disambiguation module.
Firstly, taking the reranked passages and the prompt question $Q^p$ as input, the {Answer Prediction} module generates one or multiple plausible answers $A_1, ..., A_m$.
If multiple plausible answers are found, the prompt question is treated as ambiguous so that the {Question Disambiguation} module generates a disambiguated question $Q^d_i$ for each predicted answer $A_i$.
Note that our general pipeline in Figure \ref{fig:model} does not limit the implementation of Answer Prediction module and Question Disambiguation module, and it can work for our \modelnameshort\ as well as several baselines (shown in Sec. \ref{sec:experi-round}).
Our implementation is detailed in Sec. \ref{sec:single-detail}.

\begin{figure*}[t!]
\centering
\begin{subfigure}[b]{.45\linewidth}
\includegraphics[width=\linewidth]{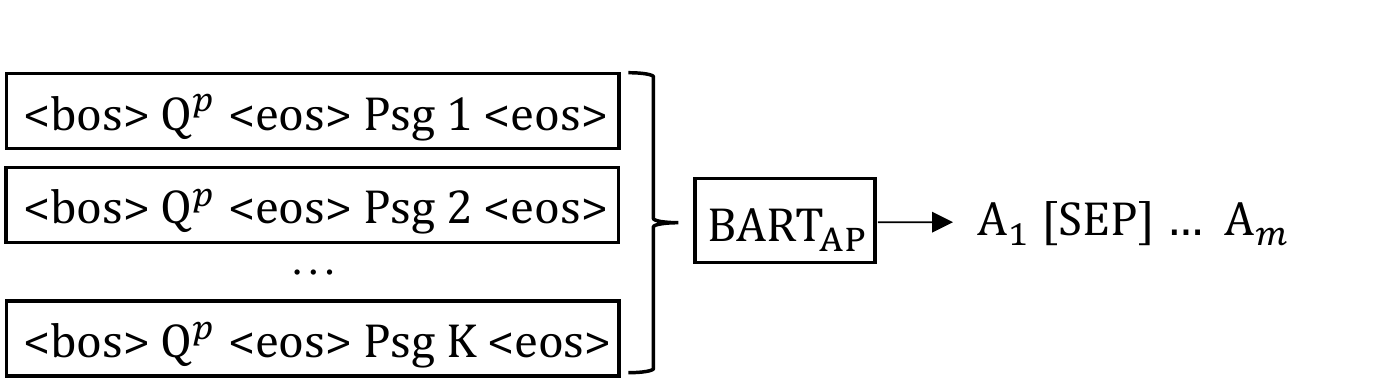}
\caption{Answer Prediction Module}\label{fig:ap}
\end{subfigure}
\begin{subfigure}[b]{.45\linewidth}
\includegraphics[width=\linewidth]{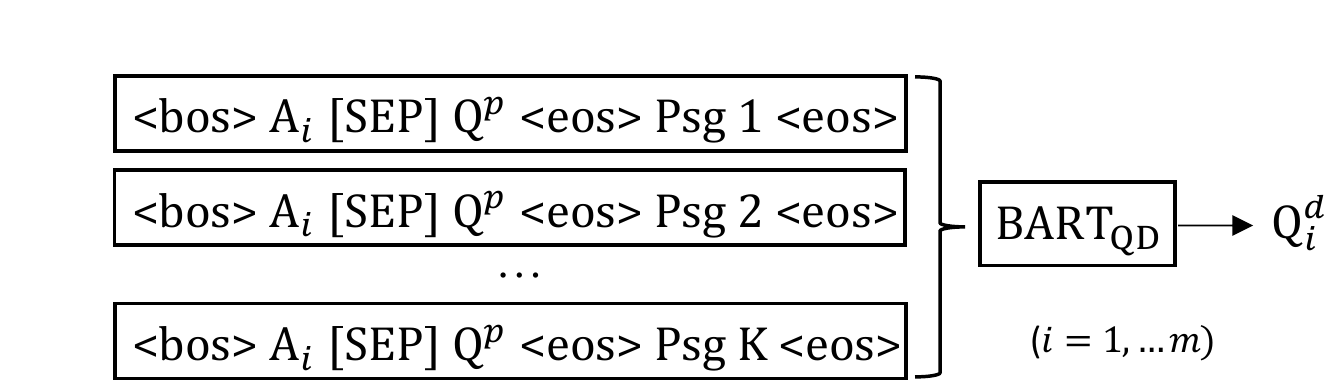}
\caption{Question Disambiguation Module}\label{fig:qd}
\end{subfigure}
\caption{The architecture of single pass QA pair generation in \modelnameshort.}
\label{fig:qagen}
\end{figure*}

\subsection{Round-Trip Prediction}\label{sec:round}
During answering ambiguous questions, it might be difficult to find every possible interpretation in the first prediction pass, and existing work \cite{min-etal-2020-ambigqa} predicts 47\% less answers compared with the ground truth.
Therefore, we propose \textit{round-trip prediction}, which includes a Round-Trip Generation step and a Language Model Verification Step. 

\paragraph{Round-Trip Generation.}
Keeping the same retrieved passages, we continuously feed the generated disambiguated questions into the Answer Prediction module to check if any new answers are generated, and generate their corresponding disambiguated questions until there are no newly predicted answers.
As exemplified in Figure \ref{fig:model}, $(\textrm{Q}_1^d, \textrm{A}_1), (\textrm{Q}_2^d, \textrm{A}_2)$ are two disambiguated QA pairs of the ambiguous prompt question $Q^p$ after the first prediction pass. When feeding $\textrm{Q}_1^d$ to the Answer Prediction module again ($1^{\text{st}}$ Round-Trip Prediction), we find that besides the previously predicted answer $\textrm{A}_1$, a new answer candidate $\textrm{A}_3$ is predicted. 
Then we generate its corresponding question $\textrm{Q}_3^d$ accordingly.
This loop continues until there are no newly predicted answers.

\paragraph{Language Model Verification.}
Through the Round-Trip Generation, we generate a bunch of QA pairs from the ambiguous prompt question, but some of them are incorrect.
Here we adopt a verification process to filter out these incorrect predictions.
Recent works in synthetic QA pair generation \cite{alberti-etal-2019-synthetic,Puri2020TrainingQA} use an ``Exact Match (EM) Verification'' approach to prune the QA pairs.
They separately train a QA model as the verification model, and drop the predicted $(q,a)$ when the verification model's answer $a' \neq a$.
However, this EM Verification approach is only suitable for factoid reading comprehension tasks such as SQuAD \cite{rajpurkar-etal-2016-squad}, in which the QA model has near-human accuracy so that it will not falsely filter out too many correct QA pairs.
In open-domain QA, the current best model can only have 51.4\% EM accuracy on the \nqopen\ dataset \cite{Izacard2020LeveragingPR}.

Instead of using hard filtering, we employ a ``Language Model (LM) Verification'' approach that is similar to the \emph{LM filtering} method of \citet{shakeri-2020}.
LM Verification is a conditional-probability-based approach to filter out QA pairs softly.
In ``LM Verification'', we first train a conditional language model using the gold disambiguated QA pairs from \ambignq.
The conditional language model is trained to estimate the likelihood of an answer given the golden disambiguated question.
Once training is done, it is used to score the generated QA pair $(q, a)$ from \modelnameshort, which is the likelihood of the answer $a$ given the question $q$ and passages,
\begin{align}
    \textrm{LM score} = \Sigma_{i=1}^{N_a} \textrm{log}~p(a^i|q, \textrm{passages}),
\end{align}
where $N_a$ is the length of the generated answer.
Finally, we rerank all predicted QA pairs according to the LM score, and drop the QA pairs according to a threshold $\text{Th}=6.1$.
The threshold is tuned according using the development set.

\section{Single Pass QA Pair Generation Details}\label{sec:single-detail}

\subsection{Answer Prediction}\label{sec:ap}

\spanseqgen\ \cite{min-etal-2020-ambigqa} concatenates the prompt question and top reranked passages into a single sequence for BART encoding, which is extremely limited by the maximum input sequence length of BART (1024 subwords, equivalent to 8 passages).
Consequently, \spanseqgen\ finds fewer interpretations of the prompt question compared to the ground truth (1.17 vs 2.19).
To ensure a broad coverage of retrieved \& reranked passages, our {Answer Prediction} module uses the Fusion-in-Decoder approach \cite{Izacard2020LeveragingPR}, which allows us to scale the number of processed passages.
As shown in Figure \ref{fig:qagen}, our BART-based Answer Prediction module $\textrm{BART}_\textrm{AP}$ encodes the concatenation of the prompt question and each passage independently. 
Then all encoded token-level representations are concatenated into a single sequence, and the $\textrm{BART}_\textrm{AP}$ decoder performs attention over all passages to aggregate and combine evidence.
Finally, the $\textrm{BART}_\textrm{AP}$ decoder generates a sequence of plausible answers token-by-token, separated by \texttt{[SEP]}.
Since there is no cross-passage attention in the encoder, $\textrm{BART}_\textrm{AP}$ encoder reduces the computation from quadratic in the number of input passages to linear complexity.
As a result, it can process 12 times larger number of input passages (up to 100 passages, 16000 subwords) than \spanseqgen.
Given that \ambignq\ is a small dataset with only 10k training samples, we first pre-train $\textrm{BART}_\textrm{AP}$ on \nqopen\ to predict a single answer, then fine-tune it on \ambignq\ to predict one or multiple answers.

\subsection{Question Disambiguation}\label{sec:qd}
If multiple answers are predicted, the Question Disambiguation module is activated to generate a disambiguated rewriting of the prompt question for each predicted answer.
Because we do not know which input passage is the key evidence to derive the predicted answer, the Question Disambiguation module takes the same passages in the Answer Prediction stage as inputs.
Similar to the Answer Prediction module $\textrm{BART}_\textrm{AP}$, our Question Disambiguation module $\textrm{BART}_\textrm{QD}$ processes the inputs under the same fashion except that $\textrm{BART}_\textrm{QD}$ encoder additionally takes the predicted answer $\textrm{A}_i$ from $\textrm{BART}_\textrm{AP}$ in the input (shown in Figure \ref{fig:qagen}).

\paragraph{Token-Deletion Pre-training.}
Similar to the training scheme of the Answer Prediction module, we also want to leverage the large-scale \nqopen\ data for pre-training.
One straightforward way is to train a question generation model on \nqopen\ that generates questions given the passages and answer, and then fine-tune it for question disambiguation on \ambigqa\ given the prompt question, answer, and passages.
However, there is no input question to disambiguate in the question generation pre-training task, it leads to a mismatch between pre-training and fine-tuning. Ablation study shows this way of pre-training has almost no help for question disambiguation (Section \ref{sec:ablation}).

To reduce the mismatch issue between pre-training and fine-tuning, we propose a Token-Deletion Pre-training task.
The idea is to construct synthetic ambiguous questions in pre-training to reduce the mismatch.
Given a question $Q$ from \nqopen, we randomly delete an \textit{informative span} from it, resulting in a partial question $Q^{s}$.
This partial question is designed to simulate the ambiguous question $Q^p$ in the fine-tuning stage.
Then the token-deletion pre-training target is to recover the complete question $Q$ from the partial question $Q^s$, answer, and passages.
In this way, the token-deletion pre-training aligns the fine-tuning phase.

Prompt questions are usually rewritten by adding new constraints including event/entity references, properties, answer types, etc. For example, the disambiguated question $\text{Q}_1$ in Figure \ref{fig:example} inserts ``by a combined team'' after the ambiguous prompt question.
Therefore, we define the \textit{informative span} as the span containing at least one of the following Part-of-Speech tags: 'ADJ', 'NOUN', 'NUM', 'PROPN', 'SYM', 'VERB'. 
The length of the span is uniformly sampled in $[1,5]$.

\paragraph{Insertion-based Weighted Loss.}
Since the disambiguated question is a small modification from the ambiguous prompt question, most tokens can be directly copied from the input.
Here we introduce an insertion-based weighted loss to put more emphasis on the newly added tokens of the disambiguated question, which could be the key to disambiguate the prompt question.
Given the prompt question $Q^p$, we find the newly inserted tokens from the disambiguated question $Q^d$: $\{q^{in}\}$.
The final loss for fine-tuning $\textrm{BART}_\textrm{QD}$ is a combination of the original negative log-likelihood loss on all question tokens augmented with a term that adds weight on the likelihood of inserted tokens:
\begin{align}\label{eqn:weighted_loss}
    \mathcal{L} = \mathcal{L}_{nll} - \lambda \sum_{q_j \in \{q^{in}\}} \textrm{log} (q_j |A, Q^p, \textrm{Psg}),
\end{align}
where $\mathcal{L}_{nll} = \sum_{i=1}^{n} \textrm{log} (q_i |A, Q^p, \textrm{Psg})$, $n$ is the number of tokens in the disambiguated question, $\lambda=3.5$ is a hyperparameter tuned on the dev. set.

\begin{table*}[!t]
	\centering
	\small
	\resizebox{0.85\textwidth}{!}{
	\begin{tabular}{lcccccccccc}
    \toprule
        \multirow{2}{*}{Model} & \multicolumn{2}{c}{\Fanswer\ (all)} &  \multicolumn{2}{c}{\Fanswer\ (multi)} & \multicolumn{2}{c}{\Fbleu} & \multicolumn{2}{c}{\Fedit} & \multicolumn{2}{c}{Comb.} \\
        \cmidrule(lr){2-3} \cmidrule(lr){4-5} \cmidrule(lr){6-7} \cmidrule(lr){8-9} \cmidrule(lr){10-11} 
        & {dev} & {test} & {dev} & {test} &{dev} & {test} & {dev} & {test} & {dev} & {test} \\
    \midrule
        \textsc{Disambig-first} \cite{min-etal-2020-ambigqa} & 28.1 & 24.8 & 21.9 & 18.8 & 4.2 & 4.0 & 2.7 & 2.2 & 30.8 & 27.0 \\
        DPR Reader \cite{min-etal-2020-ambigqa} & 37.1 & 32.3 & 28.4 & 24.8 & 13.4 & 11.3 & 6.6 & 5.5 & 43.7 & 37.8 \\
        \spanseqgen\ \cite{min-etal-2020-ambigqa} & 39.7 & 33.5 & 29.3 & 24.5 & 13.4 & 11.4 & 7.2 & 5.8 & 46.9 & 39.3 \\
        \modelnameshort\ w/o RTP (single model) & \textbf{48.4} & 41.7 & 37.0 & 32.7 & 16.0 & 14.8 & 11.2 & 9.0 & 59.6 & 50.7 \\ 
        \modelnameshort\ (single model) & {48.3} & \textbf{42.1} & \textbf{37.3} & \textbf{33.3} & \textbf{16.2} & \textbf{15.3} & \textbf{11.8} & \textbf{9.6} & \textbf{60.1} & \textbf{51.7} \\ 
    \midrule
        \spanseqgen\ (ensemble) & 41.2 & 35.2 & 29.8 & 24.5 & 13.6 & 10.6 & 7.4 & 5.7 & 48.6 & 40.9 \\
        \modelnameshort\ (ensemble) & \textbf{50.4} & \textbf{44.3} & \textbf{38.7} & \textbf{34.8} & \textbf{17.0} & \textbf{15.9} & \textbf{12.5} & \textbf{10.1} & \textbf{62.9} & \textbf{54.4} \\
    \bottomrule
    \end{tabular}
    }
	\caption{
	   Results on the dev. and hidden test set of \ambignq. 
	   ``\modelnameshort\ w/o RTP'' is the single pass prediction model without using round-trip prediction.
	   In addition to metrics introduced in Section \ref{sec:exp-setup}, we also show a combined metric ``Comb.'' = \Fanswer\ (all) + \Fedit\, which is used to rank models on the official leaderboard.
	}\label{tab:main}
\end{table*}

\section{Experiments}

\subsection{Experimental Setup}\label{sec:exp-setup}

\paragraph{Dataset.}
We conduct main experiments on the \ambignq\ dataset \cite{min-etal-2020-ambigqa}. 
\ambignq\ is constructed to address the ambiguity of questions in open-domain QA.
It samples 14,042 questions from \nqopen, a large-scale open-domain QA dataset in which each question has a single answer \cite{kwiatkowski2019natural}, and asks annotators to search for, navigate and read multiple Wikipedia pages to find as many interpretations as possible.
As a result, each question is annotated with either a single answer or multiple disambiguated QA pairs, depending on how many interpretations can be found.
The train, development, and test (not public) dataset sizes are 10036, 2002, 2004, respectively
\footnote{Leaderboard: \url{https://nlp.cs.washington.edu/ambigqa/leaderboard.html}}.
On average, there are 2.1 distinct answers per question in \ambignq. 
To test the generalization ability of \modelnameshort\ on any possibly ambiguous questions, we additionally evaluate it on two open-domain QA datasets: \nqopen\ and \triviaqa\ \cite{joshi-etal-2017-triviaqa}.

\paragraph{Implementation Details} are in Appendix \ref{sec:implementation}. We release source code for our models and experiments at \url{https://github.com/amzn/refuel-open-domain-qa}.

\paragraph{Evaluation Metrics.}
Let $(q_1,a_1), ..., (q_m, a_m)$ be $m$ QA pair predictions, $(\hat{q}_1,\hat{a}_1)$, ..., $(\hat{q}_n,\hat{a}_n)$ be $n$ gold QA pairs, each predicted QA pair $(q_i,a_i)$ is evaluated in order by a \textit{correctness score} towards all gold QA pairs:
$c_i=\mathbbm{1}(a_i\textrm{=}\hat{a}_j)f(q_i, \hat{q}_j)$,
where $f(q_i, \hat{q}_j)$ is a similarity function for questions.
$(\hat{q}_j,\hat{a}_j)$ will not be further used to evaluate other predicted QA pairs as it is used for $(q_i,a_i)$.
The overall correctness is calculated by F1 between predictions and references,
\begin{align}
    \textrm{P}_f = \frac{\sum_{i=1}^{m} c_i}{m}, \textrm{R}_f = \frac{\sum_{i=1}^{m} c_i}{n}, \textrm{F1}_f = \frac{2\textrm{P}_f\textrm{R}_f}{\textrm{P}_f+\textrm{R}_f}. \nonumber
\end{align}
\noindent All examples are evaluated for the answer prediction subtask, in which $f$ function always yields 1. This metric is denoted as \Fanswer\ (all).
For the subset of examples with multiple \textbf{gold} QA pairs, both answer prediction subtask and question disambiguation subtask are evaluated.
The answer prediction metric only computed on this subset is denoted as \Fanswer\ (multi). 
To evaluate question disambiguation performance, BLEU \cite{papineni-etal-2002-bleu} and \textsc{EDIT}-F1 is used for the function $f$, denoted as \Fbleu\ and \Fedit, respectively.
\textsc{EDIT}-F1 compute the F1 score of added and deleted unigrams from the prompt question to the predicted disambiguated question towards references.

\subsection{Experimental Results} \label{sec:results}
\paragraph{Main Results.}
Performance on the dev. and hidden test set of \ambignq\ is shown in Table \ref{tab:main}.
Even without having round-trip prediction, \modelnameshort\ (w/o RTP) outperforms \spanseqgen\ on both the answer prediction subtask and question disambiguation subtask by a large margin.
Moreover, the round-trip prediction indeed further improves the performance by finding more and better QA pairs, going from 1.55 to 1.72 pairs per prompt question on the dev. set.
A comprehensive analysis on the round-trip prediction is discussed in Sec \ref{sec:experi-round}.

\begin{table}[!t]
    \small
    \centering
    \resizebox{1.0\columnwidth}{!}{
    \begin{tabular}{l | c c c c c }
    \toprule
    Model & N & K  & \#QAs & \Fanswer\ & \Fedit\ \\
    \midrule
    \spanseqgen\    & 100 & $\approx$8 & 1.17 & 39.7 & 7.2       \\
    \spanseqgen*    & 100 & $\approx$8 & 1.14 & 41.7 & 7.1        \\
    \modelnameshort\ (w/o RTP) & 100 & 8 & 1.42 & 44.7 & 10.0      \\
    \modelnameshort\ (w/o RTP) & 100 & 100 & 1.54 & 45.4 & 10.7      \\
    \modelnameshort\ (w/o RTP) & 1000 & 100 & 1.55 & 48.4 & 11.2     \\
    \bottomrule
    \end{tabular}
    }
    \caption{
    Dev. set results of \ambigqa\ as a function of the number of retrieval/reranking (N) and QA input (K) passages. \#QAs: the average number of predicted QA pairs per prompt question. *: our replicated results.
    }
    \label{tab:ablation_psgs_input}
\end{table}

\paragraph{Controlled Comparison with \spanseqgen.}
Besides round-trip prediction, \modelnameshort\ has two advantages over \spanseqgen\ in terms of input passages:
(1) We retrieve top N=1000 passages (instead of 100 in \spanseqgen) to get a higher answer recall at top 100 passages (improved from 86.2 to 89.7).
(2) \modelnameshort\ takes K=100 input passages whereas \spanseqgen\ takes at most 1024 subwords (K$\approx$8).
To establish a controlled and fair comparison, we remove the round-trip prediction part of \modelnameshort, and feed \modelnameshort\ (w/o RTP) with the same input passages used in \spanseqgen\ (N=100, K=8).
Results are shown in Table \ref{tab:ablation_psgs_input}. We find
(1) Under the same number of passages, \modelnameshort\ (w/o RTP) (N=100, K=8) still outperforms \spanseqgen\ and generates more and better QA pairs;
(2) \modelnameshort\ (w/o RTP) benefits from increasing the answer recall of retrieval stage ($\text{N}=100 \rightarrow 1000$), as well as allowing more input passages ($\text{K}=8 \rightarrow 100$).

\begin{table}[!t]
    \small
    \centering
    \resizebox{1.0\columnwidth}{!}{
    \begin{tabular}{l c c c c }
    \toprule
    \multirow{3}{*}{Model} & \multicolumn{2}{c}{\nqopen} & \multicolumn{2}{c}{\triviaqa} \\
    \cmidrule(lr){2-3} \cmidrule(lr){4-5} 
    & {EM} & {Oracle EM} & {EM} & {Oracle EM}  \\
    \midrule
    ORQA (supervised) & 33.3 & - & 45.0 & - \\
    HardEM (supervised) & 28.1 & - & 50.9 & - \\
    DPR (supervised) & 41.5 & - & 57.9  & -       \\
    RAG (supervised)  & 44.5 & - & 56.8  & - \\
    \midrule
    \modelnameshort\ w/o RTP (NFT) & 35.4 & 45.2 & 48.2 & 52.9      \\
    \modelnameshort\ (NFT) & 37.3 & 48.9 & 49.8 & 54.3      \\
    \bottomrule
    \end{tabular}
    }
    \caption{Results on \nqopen\ and \triviaqa\ test set. 
    RTP: Round-Trip Prediction. NFT: No Fine-Tuning.
    ORQA \cite{lee-etal-2019-latent}, HardEM \cite{min-etal-2019-discrete}, RAG \cite{Lewis2020RetrievalAugmentedGF}.
    }
    \label{tab:other_datasets}
\end{table}

\begin{table*}[!t]
    \small
    \centering
    \resizebox{1.0\textwidth}{!}{
    \begin{tabular}{l | c c c c c c c}
    \toprule
    Models & \#QAs & \Fanswer\ (all) & \Fanswer\ (multi) & \Fbleu\ & \Fedit\ & Comb.  \\
    \midrule
    \modelnameshort\ w/o RTP    & 1.55 & {48.4}~~ & 37.0~~ & 16.0~~ & 11.2~~ & 59.6   \\
    ~~~+ Round-Trip Generation  & 2.06 ($\uparrow$33.5\%) & 47.6~~ & {37.4}~~ & 16.0~~ & 11.4~~ & 59.0~~ ($\downarrow$0.9\%) \\
    ~~~+ Round-Trip Generation \& LM Verification   & 1.72 ($\uparrow$11.1\%) & 48.3~~ & 37.3~~ & {16.2}~~ & {11.8}* & {60.1}~~ ($\uparrow$0.7\%)  \\
    ~~~+ Round-Trip Generation \& EM Verification     & 1.43 ($\downarrow$~~7.7\%)  & 47.6~~ & 35.4~~ & 15.7~~ & 11.6~~ & 57.2~~ ($\downarrow$4.0\%) \\
    \midrule
    DPR Reader & 1.62 & 38.9~~ & 29.9~~ & 12.5~~ & 6.8~~ & 45.7 \\
    ~~~+ Round-Trip Generation \& LM Verification & 1.81 ($\uparrow$11.7\%) & 40.1* & 	31.6* & 	13.3* & 	7.3* & 47.4* ($\uparrow$3.7\%) \\
    \spanseqgen\ & 1.14 & 	41.7~~ & 	29.3~~ & 	12.7~~ & 	7.1~~ & 	48.8 \\
    ~~~+ Round-Trip Generation \& LM Verification & 1.28 ($\uparrow$12.3\%) & 	42.4* & 	29.9* & 	13.0* & 	7.4* & 49.8* ($\uparrow$2.1\%) \\
    \bottomrule
    \end{tabular}
    }
    \vspace{-0.1in}
    \caption{
    Effect of \textit{round-trip prediction} to harvest more interpretations (QA pairs) on the development set of \ambigqa. 
    ``$\uparrow \text{and} \downarrow$'' denotes the improvement gain over the model without round-trip prediction.
    *: The model with "Round-Trip Generation \& LM Verification" is significantly better than the same model without it under a paired bootstrap test with ${10}^5$ samples (\textit{p-value} \textless 0.05).
    }
    \label{tab:roundtrip}
\end{table*}

\paragraph{Generalization to Other Datasets.}
To test how well does \modelnameshort\ answer any open-domain questions, we evaluate \modelnameshort\ on \nqopen\ and \triviaqa\ without finetuning on these datasets.
When \modelnameshort\ predicts multiple answers, we take the first predicted answer for EM evaluation; we also introduce a new \textbf{Oracle EM} metric which treat the prediction is correct if the gold answer matches any predicted answers for the current question.
Table \ref{tab:other_datasets} shows that \modelnameshort\ has competitive performance even without dataset-specific finetuning.
When \modelnameshort\ finds multiple interpretations for questions in \nqopen\ \& \triviaqa, we manually check the quality of disambiguated QA pairs in Section \ref{sec:human_evaluation}.

\subsection{Effect of Round-Trip Prediction} \label{sec:experi-round}

We compare our proposed Round-Trip Prediction (Round-Trip Prediction = Round-Trip Generation + LM Verification) with several alternative approaches, as well as investigate its generalization ability to other models like \spanseqgen\ and DPR Reader.
Results are shown in Table \ref{tab:roundtrip}.

\paragraph{Round-Trip Generation Only.}
We investigate the necessity of the verification process by conducting only round-trip generation to \modelnameshort.
Results show that Round-Trip Generation can generate 33.5\% more QA pairs, but the lower \Fanswer\ (all) suggests that this strategy may over-generate QA pairs when the prompt question is not ambiguous.
Hence, the verification process is necessary to prune some incorrect QAs.

\paragraph{LM Verification \textit{vs.} EM Verification.}
As described in section \ref{sec:round}, we compare the existing EM Verification approach \cite{alberti-etal-2019-synthetic, Puri2020TrainingQA} with our LM Verification.
Results demonstrate that EM Verification prunes too many QA pairs -- the number of remaining QA pairs (1.43) is even smaller than not doing round-trip prediction (1.55).
This validates our intuition in section \ref{sec:round} that EM Verification is not suitable for open-domain QA tasks because of the low performance of current open-domain QA models.

\paragraph{Generalization to Other Models.}

We show that round-trip prediction is a model-agnostic general approach for answering possibly ambiguous open-domain questions by using it on our replicated baseline models: DPR Reader and \spanseqgen.
With the help of round-trip prediction, DPR Reader and \spanseqgen\ generates 11.7\% and 12.3\% more QA pairs, which result in a boost of 3.7\% and 2.1\% for the overall performance (Comb.).

\begin{table}[!t]
    \small
    \centering
    \resizebox{1.0\columnwidth}{!}{
    \begin{tabular}{l | c c c c c}
    \toprule
    Models & Dataset & \#QAs & \#C-QAs & \#CD-QAs & $\kappa$ \\
    \midrule
    \spanseqgen\ & \ambigqa & 2.12 & 1.40 & 0.46 & 0.27  \\
    \modelnameshort\ w/o RTP & \ambigqa & 2.80 & 1.84 & 0.98 & 0.35 \\
    \modelnameshort\  & \ambigqa & 3.44 & 2.40 & 1.24 & 0.34 \\
    \midrule
    \modelnameshort\ w/o RTP  & \nqopen & 2.32 & 1.30 & 0.64 & 0.20 \\
    \modelnameshort\  & \nqopen & 3.20 & 1.72 & 0.88 & 0.21 \\
    \midrule
    \modelnameshort\ w/o RTP  & \triviaqa & 2.08 & 1.02 & 0.46 & 0.34 \\
    \modelnameshort\  & \triviaqa & 3.24 & 1.84 & 0.82 & 0.35 \\
    \bottomrule
    \end{tabular}
    }
    \caption{
    Human evaluation results.
    \#QAs: the average number of QA pairs per prompt question.
    \#C-QAs \& \#CD-QAs: the average number of \textit{correct} QA pairs, detailed in Sec. \ref{sec:human_evaluation}.
    $\kappa$: Fleiss' kappa score.
    }
    \label{tab:human_eval}
\end{table}

\subsection{Human Evaluation} \label{sec:human_evaluation}
Since the answers collected in \ambigqa\ are not necessarily exhaustive, there is a possibility that a model generates correct interpretations but they are missed in \ambigqa.
Therefore, we hire 3 workers from \url{MTurk.com} to evaluate the correctness of the answer given the generated disambiguated question and retrieved passages (instructions in Appendix \ref{sec:detail_human_eval}).
Let $(q_1, a_1), ..., (q_n, a_n)$ be $n$ generated QA pairs from the same prompt question, we define two levels of correctness as follows:
\textbf{\#C-QAs}: $(q_i, a_i)$ is considered \textbf{C}orrect if $a_i$ is a correct answer of $q_i$; 
\textbf{\#CD-QAs}: $(q_i, a_i)$ is considered correct iff. (1) $a_i$ is a correct answer of $q_i$ \textit{and} (2) any $a_j (j \neq i)$ is a wrong answer of $q_i$.
\#CD-QAs is designed to examine the \textbf{C}orrectness of question \textbf{D}isambiguation because ambiguous questions can have multiple valid answers. 
We take the majority judgement from 3 annotators for each QA pair.
For each dataset, we randomly sample 50 prompt questions which have multiple predicted answers, and apply the QA swapping strategy in \#CD-QAs, resulting 960 question-answer-passages triples in total.
Results in Table \ref{tab:human_eval} show that \modelnameshort\ (w/o RTP) can correctly generate 113\% more QA pairs than \spanseqgen\ on \#CD-QAs.
In addition, round-trip prediction (RTP) can find more correct interpretations across all datasets.

\subsection{Ablations on Question Disambiguation}\label{sec:ablation}

\begin{table}[!t]
    \small
    \centering
    \resizebox{1.0\columnwidth}{!}{
    \begin{tabular}{l | c c }
    \toprule
    Pre-train Method + Fine-tune Method & \Fbleu\ & \Fedit\ \\
    \midrule
    Prompt Baseline & 18.9 & 0.0  \\
    None + \QDF\ & 16.2 & 10.1 \\
    None + \QDF\ (w/ filtered passages)  & 16.4 & 9.4 \\
    \QGP\ + \QDF\  & 15.9 & 10.3 \\
    \TDP\ + \QDF\  & 16.5 & 10.9 \\
    \TDP\ + \QDF\ (w/ insertion-based loss)  & 16.0 & \textbf{11.2}   \\
    \bottomrule
    \end{tabular}
    }
    \caption{
    Ablation Study of \modelnameshort\ for the question disambiguation subtask on the dev. set. \QDF: Question Disambiguation Fine-tuning, \QGP: Question Generation Pre-training, \TDP: Token-Deletion Pre-training.
    }
    \label{tab:ablation_qd}
\end{table}

Table \ref{tab:ablation_qd} compares our question disambiguation model with the prompt baseline and several ablations.
The prompt baseline directly takes the prompt question as the disambiguated prediction, so its \Fedit\ is zero. 
However, \Fbleu\ score of the prompt baseline is higher than \modelnameshort.
This suggests that \Fedit\ captures the effectiveness of question disambiguation better than \Fbleu.

For our ablations, we start from only using \ambigqa\ dataset (None+\QDF), and investigate whether it is helpful to only use answer-containing passages as inputs (None+\QDF\ w/ filtered passages).
The worse result of the latter approach suggests that we should keep all passages for question disambiguation.
Second, we examine the effectiveness of pre-training.
We try the question generation pre-training (\QGP+\QDF) and compare it with the ablation without any pre-training (None+\QDF). Results show that the question generation pre-training has little help for fine-tuning.
By replacing the question generation pre-training \QGP\ with our proposed token-deletion pre-training \TDP, we see the results (\TDP+\QDF) are better than the no pre-training ablation (None+\QDF), which implies the mismatch between pre-training and fine-tuning are somewhat reduced.
Finally, the insertion-based loss enables \modelnameshort\ to capture the key disambiguation phrase with less copying the prompt question, resulting in a lower BLEU but higher Edit-F1.

\begin{figure}[!t]
    \centering
    \footnotesize
    \setlength\tabcolsep{3.8pt}
    \begin{tabular}{p{0.95\columnwidth}}
    \toprule
        \textbf{\Prompt\ \#1:} What's the most points scored in an nba game? \\
        \textbf{Reference:} \\
        Q1: What is the highest amount of points scored by a single team in regular season NBA games? / A1: 186 \\
        Q2: What is the highest amount of points scored by a single team in regular season games in regulation? / A2: 162 \\
        Q3: What is the highest amount of points scored by a single team in playoff games? / A3: 153 \\
        \textbf{\modelname\ w/o RTP}: (QA1-QA4: \Fanswer=57.1, \Fedit=44.9) \\
        Q1: What's the most points scored in a regular season nba game by combined? / A1: 370 \\
        Q2: What's the most points scored in an nba playoff game by combined? / A2: 304 \\
        Q3: What's the most points scored in an nba game by individual? / A3: 100 \\
        Q4: What player scored the most points in an NBA game? / A4: wilt chamberlain \\
        \textbf{\modelname}: (QA1-QA6: \Fanswer=66.7, \Fedit=57.1) \\
        Q5: What's the most points scored in an NBA game by single team? / A5: 186 \\
        Q6: What's the most points scored in an nba playoff game by single team? / A6: 153 \\
        \textbf{Relevant Passages:} (w/ rank from retrieval \& reranking) \\
        \underline{Rank 1}:   ... the highest-scoring regular season game is ... the two teams combined to score \hl{370} points, with the pistons defeating the nuggets \hl{186}–184 ... \\
        \underline{Rank 3}:  \hl{wilt chamberlain} scored an nba-record \hl{100} points. the highest-scoring playoff game is the double-overtime game between ... the two teams combined to score \hl{304} points, with the trail blazers defeating the suns \hl{153}–151 ... \\
    \bottomrule
    \end{tabular}
    \caption{
        Predictions generated by \modelname\ w/o round-trip prediction (QA1-QA4) and \modelname\ (QA1-QA6). 
    }
    \label{tab:case}
\end{figure}

\subsection{Case Study}
Figure \ref{tab:case} provides example question-answer pairs generated by crowd-workers, \modelnameshort\ (w/o RTP), and \modelnameshort. 
The annotator find three interpretations from the prompt question, while our single pass model \modelnameshort\ (w/o RTP) finds in total four interpretations (QA1-4).
Although QA2 predicted from our model is not included in the references, it is indeed a correct interpretation of the prompt question.
In addition, the Round-Trip Prediction approach finds two correct interpretations (QA5, QA6) which the model fails to predict on the first generation pass. 
More cases are shown in Appendix \ref{sec:more_cases}.

\section{Related Work}
Open-Domain Question Answering is answering factoid questions using a huge collection of documents such as Wikipedia pages \cite{Voorhees1999TheTQ, chen-etal-2017-reading,yang-etal-2019-end-end-open,lee-etal-2019-latent,wang-etal-2019-multi}. 
We are motivated by the recent proposed question ambiguity problem in open-domain QA \cite{min-etal-2020-ambigqa}.
Different from the existing formulation of open-domain QA that  each question only has a single answer, the proposed \ambigqa\ task requires to predict a single answer or a set of disambiguated QA pairs depending on the ambiguity of the input question.
They also propose the first model \spanseqgen\ to this task, which firstly uses the dense passage retriever \cite{karpukhin2020dense} to retrieve question-relevant passages, and then adopts a retrieval-augmented generation method \cite{Lewis2020RetrievalAugmentedGF} to disambiguated QA pairs.

Our \modelnameshort\ follow \newcite{min-etal-2020-ambigqa}'s task formulation and overall pipeline, but there are three differences between our \modelnameshort\ and \spanseqgen:
(1) \modelnameshort\ takes the architecture of Fusion-in-Decoder \cite{Izacard2020LeveragingPR} that can effectively use a large number of passages to uncover more candidate interpretations of the ambiguous question.
(2) We propose a token-deletion pre-training task to reduce the mismatch between pre-training and fine-tuning for question disambiguation. The insertion-based weighted loss further helps to capture answer-relevant constraints.
(3) We propose a model-agnostic round-trip prediction approach to find more interpretations missed in the first prediction pass, which we further refine using a conditional-probability-based filtering approach.

\section{Conclusion}
In this paper, we present \modelnameshort\ to answer ambiguous open-domain questions.
\modelnameshort\ is a generative approach to aggregate and combine evidence from multiple passages for multiple rounds which can find more and better interpretations.
\modelnameshort\ achieves a new \sota\ on \ambignq, and shows competitive performance on \nqopen\ and \triviaqa.
The proposed round-trip prediction is a general approach for answering ambiguous open-domain questions, which improves our \modelnameshort\ as well as several baseline models.




\bibliographystyle{acl_natbib}
\bibliography{anthology,acl2021}

\clearpage

\appendix

\section{Implementation Details}\label{sec:implementation}
\paragraph{Evidence Corpus.}
We keep the version of English Wikipedia Dump consistent to the annotation timestep of \nqopen\ and \ambignq, which is 2018-12-20 and 2020-01-20 respectively. Models pre-trained on \nqopen\ use passages from dump 2018-12-20 while models fine-tuned on \ambignq\ take dump 2020-01-20.
We use the \ambigqa\ processed passages of these dumps, which takes the plain text and split Wikipedia pages into 100-word passages.
As a result, there are 22M passages of Wikipedia Dump 2018-12-20 and 24M passages of Wikipedia Dump 2020-01-20.

\paragraph{Retrieval \& Reranking.} 
We use the multiset version of Dense Passage Retriever (DPR) \cite{karpukhin2020dense}, which is jointly trained on five open-domain QA datasets.
For the reranker, we fine-tune a \texttt{bert-large-cased} model with a batch size 16, learning rate 1e-5, training epoch 10 on the \nqopen\ dataset. 
We sample 1 positive and 31 negative passages in training to maximize log-likelihood of the positive passage.
The best reranker model is selected according to the answer recall in top 100 reranked passages.
The trained reranker model is used for both \nqopen\ and \ambigqa\ dataset (we tried to finetune this model on \ambigqa\ but did not receive any sensible improvement).
The total training takes 10 hours and we tune the learning rate from 1e-5 to 5e-5 and select the best one.

\paragraph{Answer Prediction.}
We train a $\textrm{BART}_\textrm{large}$ model on \nqopen\ with a batch size 64, epoch 10, and learning rate 5e-5. 
Then we finetune the trained model on \ambigqa\ with a batch size 64, epoch 30, and learning rate 3e-5. 
According to empirical results, we discard training samples which the gold answers do not appear in any input passages for training on both \nqopen\ and \ambigqa\ (in the case of \ambigqa, we discard training examples only when none of gold answers are found).
All models are selected according to the performance (EM for \nqopen, \Fanswer\ (all) for \ambignq) on the development set.

\paragraph{Question Disambiguation.}
We train a $\textrm{BART}_\textrm{large}$ model on \nqopen\ with a batch size 64, epoch 10, and learning rate 1e-5. 
Then we finetune the trained model on \ambignq\ with a batch size 64, epoch 30, and learning rate 5e-5. 
Different from training in answer prediction, we do not filter training samples which the answer does not appear in any input passages according to empirical results.
The best model is selected according to \Fedit\ for both \nqopen\ and \ambignq\ on the development set.

\paragraph{LM Verification.}
Based on the best QA model on \nqopen\ trained in the Answer Prediction, we finetune it using the gold \textit{disambiguated} QA pairs from \ambignq, in which each disambiguated question is only paired with one answer. We use a batch size 64, epoch 30, and learning rate 3e-5 for finetuning, and select the best model according to the EM score on the dev. set of \ambigqa.

All the experiments are conducted on a single machine with 8 V100 GPUs. The pre-training on \nqopen\ takes 60 hours for models in Answer Prediction, Question Disambiguation and LM Verification, and the fine-tuning takes 10 hours on \ambignq.

\begin{figure*}[!t]
    \centering
    \footnotesize
    \resizebox{1.0\textwidth}{!}{
    \setlength\tabcolsep{0.8pt}
    \begin{tabular}{p{0.15\textwidth}p{0.05\textwidth}p{0.80\textwidth}}
    \toprule
        Error Type & \% & Example \\
    \midrule
        \multirow{1.5}{*}{Wrong} &
        \multirow{3}{*}{42} &
        {Prompt Q}: How long do contestants get to answer on jeopardy? \ {Answer}: 30 seconds \\
        \multirow{1.5}{*}{Disambiguation} & & {Disamb. Q} (g): How long do contestants have to answer \hl{during the last round of} Jeopardy!? \\
         & & {Disamb. Q} (p): How long do contestants get to answer \hl{on the electronic display} on jeopardy? \\ 
    \specialrule{0em}{1pt}{1pt}
    \hdashline
    \specialrule{0em}{1pt}{1pt}
        Correct but  & \multirow{3}{*}{19} & 
        {Prompt Q}: Who is the administrator of the small business administration? \ {Answer}: mickey thomas \\
         Different & & {Disamb. Q} (g): Who is the administrator of the small business administration \hl{from 2014 to 2017}? \\
        Constraints & & {Disamb. Q} (p): Who is \hl{the 24th} administrator of the small business administration? \\
    \specialrule{0em}{1pt}{1pt}
    \hdashline
    \specialrule{0em}{1pt}{1pt}
        \multirow{1.5}{*}{Correct but} & \multirow{3}{*}{13} & {Prompt Q}: Who are the kane county cougars affiliated with? \ {Answer}: arizona diamondbacks \\
        \multirow{1.5}{*}{Paraphrase} & & {Disamb. Q} (g): Who have the Kane County Cougars been affiliated with \hl{since 2015}? \\
        & & {Disamb. Q} (p): Who are the Kane County Cougars affiliated with \hl{from 2015-present}? \\
    \specialrule{0em}{1pt}{1pt}
    \hdashline
    \specialrule{0em}{1pt}{1pt}
        \multirow{1.5}{*}{Annotation} & \multirow{3}{*}{11} & {Prompt Q}: Who played tony in only fools and horses? \ {Answer}: christopher papazoglou \\
        \multirow{1.5}{*}{Error} & & {Disamb. Q} (g): Who played tony driscoll in only fools and horses? \\
         & & {Disamb. Q} (p): Who played Tony in Only Fools and Horses \hl{from 1981-1983}? \\ 
    \specialrule{0em}{1pt}{1pt}
    \hdashline
    \specialrule{0em}{1pt}{1pt}
        \multirow{1.5}{*}{No} & \multirow{3}{*}{15} & {Prompt Q}: Who has the most nascar wins in history? \ {Answer}: richard petty \\
        \multirow{1.5}{*}{Disambiguation} & & {Disamb. Q} (g): Who has the most nascar \hl{super series} wins in \hl{all-time} history? \\
        & & {Disamb. Q} (p): Who has the most NASCAR wins in history? \\ 
    \bottomrule
    \end{tabular}
    }
    \caption{
        Types of question disambiguation errors and their proportions in the dev. data based on 100 samples. ``{Disamb. Q} (g)/(p)'': Gold/Predicted Disambiguated Question. 
        ``\textit{Correct but Different Constraints}'': Predicted questions are correct interpretations of the answers but expressed through different constraints.
        ``\textit{Correct but Paraphrase}'': Predicted questions are  paraphrases of gold questions.
        The difference between disambiguated questions and prompt questions is highlighted.
    }
    \label{tab:error}
\end{figure*}

\section{Error Analysis}\label{sec:error_analysis}

\paragraph{Answer Prediction Error.}
In the development set of \ambignq, 22.9\% of examples actually have multiple interpretations but \modelnameshort\ only predicts one answer.
In 12.0\% examples, \modelnameshort\ wrongly predicts multiple answers on the unambiguous prompt questions. 
In the rest 65.1\% examples, \modelnameshort\ aligns with annotators in terms of the ambiguity.
Since \modelnameshort\ tends to wrongly think the prompt question is unambiguous, it predicts fewer answers than ground truth (1.55 \textit{vs.} 2.02 on average). In effect, the predicted answers have a relatively high precision 55.6\% but low recall 48.0\%.
By localizing where the errors come from, we find that in 2.3\% of examples, \modelnameshort\ fails to retrieve any relevant passage which contains gold answers. In 27.0\% of examples, retrieved passages only contain part of gold answers. In 38.6\% of examples, retrieved passages can cover all gold answers but \modelnameshort\ fails to make correct predictions.

\paragraph{Question Disambiguation Error.}
We analyze the quality of disambiguated questions when the predicted answers are correct.
We select 100 samples from the development data and summarize errors into five categories in Figure \ref{tab:error}.
We see that 42\% of generated questions are totally wrong and 15\% of them are identical to the prompt ones.
Besides, there are in total 31\% of generated questions (\textit{Correct but Different Constraints}, \textit{Correct but Paraphrase}) are actually correct but do not get credits under the current matching based evaluation metric \Fedit.
This suggests that a better evaluation metric should be incorporated in future to mitigate the variability of language generation, such as using a trained QA model for evaluation.

\begin{figure*}[t!]
\includegraphics[width=0.6\textwidth]{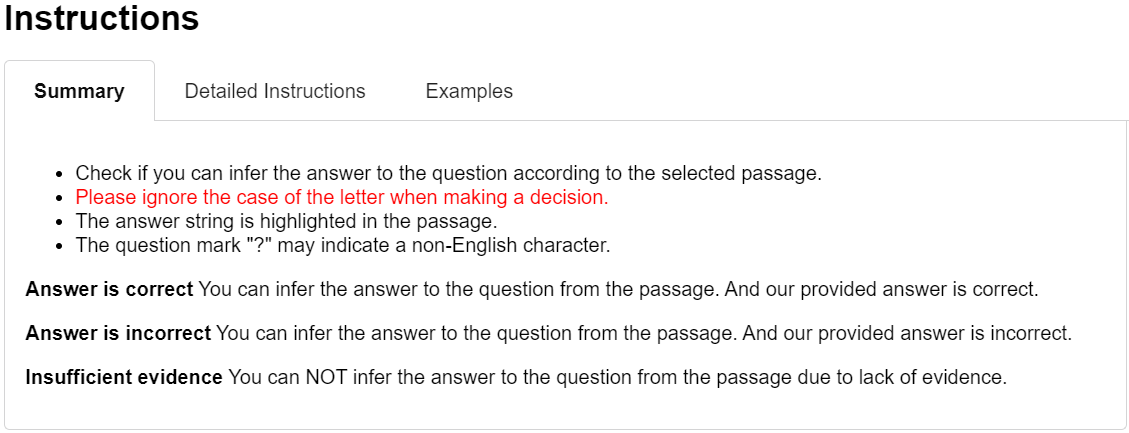}\\
\includegraphics[width=\textwidth]{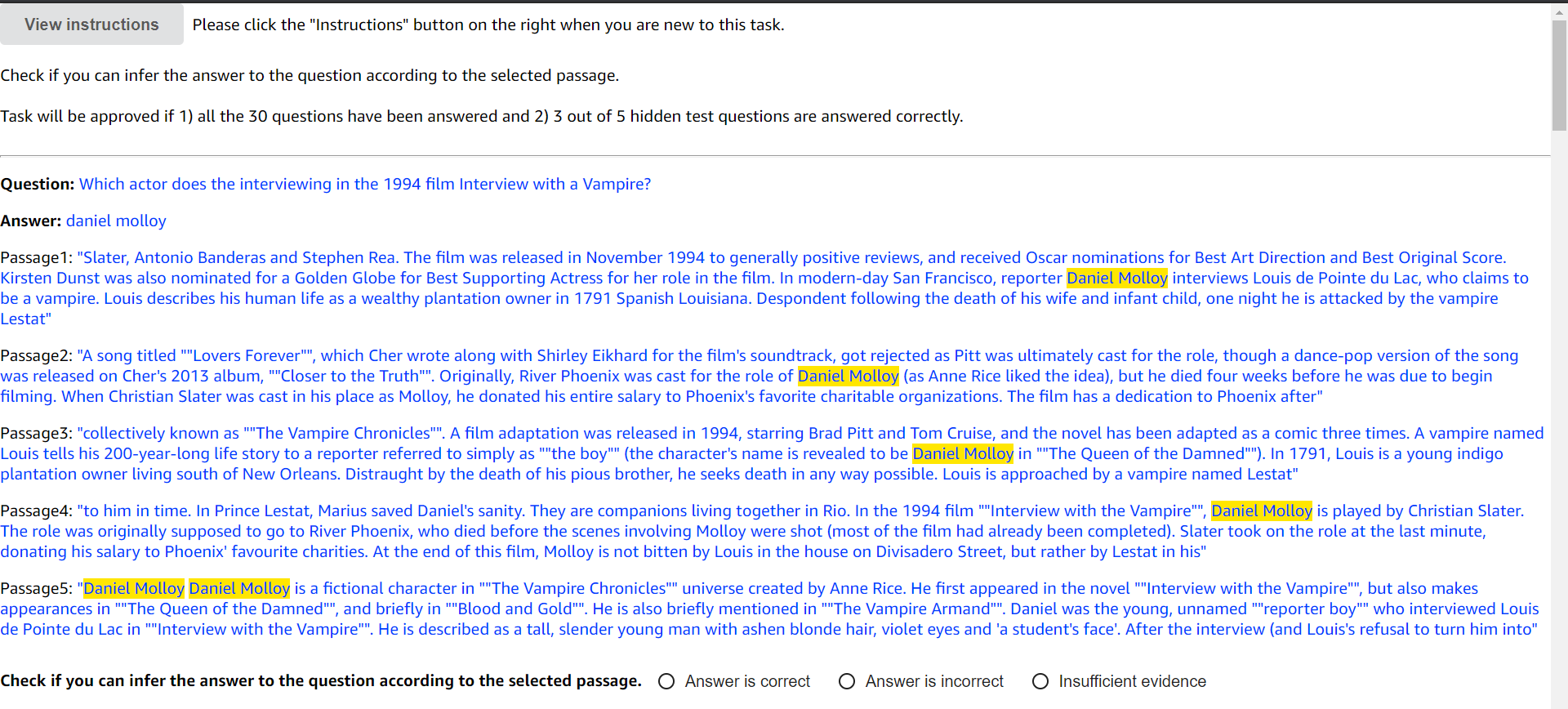}
\caption{Instructions and interface for human evaluation. \textit{(best viewed in color)}}
\label{fig:human_eval_instruction}
\end{figure*}

\section{Details of Human Evaluation} \label{sec:detail_human_eval}

\paragraph{Instruction Details.}
Figure \ref{fig:human_eval_instruction} shows the instruction and interface for human evaluation.
We have three choices for each QA pair: ``Answer is correct'', ``Answer is incorrect'' and ``Insufficient evidence''.
Since each QA pair has 100 retrieved passages, we show 5 retrieved passages (with answer highlighted) at a time.
If the worker select ``Insufficient evidence'', we will show the next 5 retrieved passages until this QA pair receives a ``correct/incorrect'' decision.
If ``Insufficient evidence'' is still select after showing all 100 passages, then we mark this QA pair as ``incorrect''.

\paragraph{Evaluation Metrics \& Quality Control.}
Let $(q_1, a_1), ..., (q_n, a_n)$ be $n$ generated QA pairs from the same prompt question, we define two levels of correctness as follows:
\textbf{\#C-QAs}: $(q_i, a_i)$ is considered \textbf{C}orrect if $a_i$ is a correct answer of $q_i$; 
\textbf{\#CD-QAs}: $(q_i, a_i)$ is considered correct iff. (1) $a_i$ is a correct answer of $q_i$ \textit{and} (2) any $a_j (j \neq i)$ is a wrong answer of $q_i$.
\#CD-QAs is designed to examine the \textbf{C}orrectness of question \textbf{D}isambiguation because ambiguous questions can have multiple valid answers. 
Moreover, it reduce the priming effect so that workers won't have a tendency to mark all samples as correct.
During annotation, workers do not know each question $q_i$ is paired with its answer $a_i$ or other answers $a_j (j \neq i)$ under the same prompt question.

We only recruit workers based in the United States and pay 0.2 USD per QA pair on Mturk. 
For quality control, we have manually annotate 15 correct QA pairs and 15 wrong QA pairs (pair $q_i$ with $a_j (j \neq i)$, and randomly select 5 of them to examine the quality of annotation. The task will be approved only when 3 out of 5 hidden test QA pairs receive correct annotations.

\section{Discussion on Problem Formulation}
\modelnameshort\ follows the problem formulation of \spanseqgen\ to firstly predict one or multiple answers, and then generate the disambiguated question for each answer.
We also tried/considered different formulations of this problem as follows:

\paragraph{QGen-AGen.} 
We swap the order of answer prediction and question disambiguation in the problem formulation -- firstly a QD model generates several disambiguated questions in a sequence, or predicts \texttt{EOS} if the question is not ambiguous; Then a QA model predicts a single answer for each predicted disambiguated question.
This approach does not work in our experiments with poor performance. We think the major reason is generating multiple disambiguated question from the prompt question as the first step is much harder than the original formulation which only requires to generating multiple plausible answers from the prompt question.

\begin{table*}[!t]
    \small
    \centering
    \resizebox{0.95\textwidth}{!}{
    \begin{tabular}{l | c c c c c c c}
    \toprule
    Models & \#QAs & \Fanswer\ (all) & \Fanswer\ (multi) & \Fbleu\ & \Fedit\ & Comb.  \\
    \midrule
    \modelnameshort\ w/o RTP    & 1.55 & \textbf{48.4} & 37.0 & 16.0 & 11.2 & 59.6   \\
    ~~~+ Round-Trip Generation  & 2.06 & 47.6 & \textbf{37.4} & 16.0 & 11.4 & 59.0 \\
    ~~~+ Round-Trip Generation \& LM Verification   & 1.72 & 48.3 & 37.3 & \textbf{16.2} & \textbf{11.8} & \textbf{60.1}  \\
    \midrule
    ~~~+ Min-Length Generation (L=8) & 2.10 & 40.8 & 36.2 & 15.5 & 11.1 & 51.9 \\
    ~~~+ Min-Length Generation (L=8) \& LM Verification & 1.69 & 42.9& 	36.3& 	15.9& 	11.4& 	54.4 \\
    ~~~+ Min-Length Generation (L=16) & 2.88 & 	37.2 & 	34.1 & 	14.6 & 	10.3 & 	47.5\\
    ~~~+ Min-Length Generation (L=16) \& LM Verification & 1.46 & 	43.1 & 	34.5 & 	15.2 & 	11.1 & 	54.2 \\
    \bottomrule
    \end{tabular}
    }
    \caption{
    Dev. set results on different approaches to harvest more interpretations (QA pairs) towards the ambiguous questions. ``\#QAs'' denotes the average number of generated QA pairs per prompt question.
    }
    \label{tab:roundtrip_baseline}
\end{table*}

\paragraph{QAGen.} 
Another possible approach is using a single model to predict disambiguated question-answer pairs where each  answer right precedes its disambiguated question. This is certainly a possible way but it is even more challenging than \textbf{QGen-AGen}. We did not try this way after receiving poor performance from \textbf{QGen-AGen}.

\section{Baselines for Round-Trip Prediction}
Since the current round-trip prediction requires several iteration between the answer prediction module and the question disambiguation module, it would be better to over-generate many answers in one pass. 
One straightforward way to generate more QA pairs is setting a minimum length of generation for the answer prediction model, and then go through the LM Verification process to drop the low-quality predictions.
We set two minimum lengths of generation (L=8/16) for our answer prediction model.
As shown in Table \ref{tab:roundtrip_baseline}, although setting a minimum length effectively increases the number of predicted QA pairs (2.10/2.88 for L=8/16), the over-generated answers are extremely noisy which in turn hurts the effectiveness of the LM Verification model, resulting in far worse performance across all metrics.
Presumably, one major disadvantage of the Min-Length Generation approach is that \modelnameshort\ loses the flexibility to decide the number of possible interpretations based on the passages. 
Instead, it always generates multiple answers according to the minimum length.


\begin{figure*}[!t]
    \centering
    \footnotesize
    \setlength\tabcolsep{3.8pt}
    \begin{tabular}{p{0.95\textwidth}}
    \toprule
        \textbf{\Prompt\ \#1:} Who played lead guitar for the rolling stones? \\
        \textbf{Reference:} \\
        Q1: Who played lead guitar for the rolling stones from 1962-1969? / A1: brian jones \\
        Q2: Who played lead guitar for the rolling stones from 1969-1974? / A2: mick taylor \\
        Q3: Who played lead guitar for the rolling stones from since 1962? / A3: keith richards \\
        Q4: Who played lead guitar for the rolling stones from since 1975? / A4: ronnie wood \\
        \textbf{Prediction of \modelname\ w/o Round-Trip Prediction:} (QA1-QA3: \Fanswer=57.1, \Fedit=8.2) \\
        Q1: Who played electric guitar for the Rolling Stones from 1962-present? / A1: keith richards \\
        Q2: Who primarily played guitar for the Rolling Stones? / A2: mick jagger and keith richards \\
        Q3: Who originally played slide guitar for the Rolling Stones? / A3: brian jones \\
        \textbf{Prediction of \modelname:} (QA1-QA4: \Fanswer=75.0, \Fedit=15.5) \\
        Q4: Who played bass guitar for the Rolling Stones from 1969-1975? / A4: mick taylor \\
        \textbf{Relevant Snippets of Passages:} (w/ rank from retrieval \& reranking module) \\
        \underline{Rank 2}: ... the original lineup consisted of multi-instrumentalist \hl{brian jones}, lead vocalist \hl{mick jagger}, guitarist \hl{keith richards}, bass guitarist bill wyman, drummer charlie watts, and keyboardist ian stewart. ... following \hl{jones}' death in 1969, \hl{mick taylor} took over lead guitar duties until 1974. \\
        \underline{Rank 4}: \hl{mick jagger} sir michael philip jagger (born 26 july 1943) is an english singer, ...  his distinctive voice and energetic live performances, along with \hl{keith richards}' guitar style, have been the trademark of the rolling stones ... \\
        \underline{Rank 10}: song, as the stones are generally known for their guitar interplay of rhythm and lead ("weaving") between \hl{richards} and the other guitarist in the band – \hl{brian jones} (1962–1969), \hl{mick taylor} (1969–1975), and \hl{ronnie wood} (1975–present) ... \\
    \midrule
        \textbf{\Prompt\ \#2:} When does the ration shop open in india? \\
        \textbf{Reference:} \\
        Q1: When did the ration shop open in india for the first time? / A1: February 1944 \\
        Q2: When did the ration shop in its current form open in india? / A2: June 1947 \\
        \textbf{Prediction of \modelname\ w/o Round-Trip Prediction:} (QA1: \Fanswer=66.7, \Fedit=28.6) \\
        Q1: When does the Indian Food Security System open in its current form? / A1: june 1947 \\
        \textbf{Prediction of \modelname:} (QA1-QA2: \Fanswer=100.0, \Fedit=71.4) \\
        Q2: When does the first ration shop open in india? / A2: february 1944 \\
        \textbf{Relevant Snippets of Passages:} (w/ rank from retrieval \& reranking module) \\
        \underline{Rank 3}: public distribution system the indian food security system was established by the government ... this scheme was first started in \hl{february 1944}, during the second world war, and was launched in the current form in \hl{june 1947}. \\
    \midrule
        \textbf{\Prompt\ \#3:} When is the new christopher robin coming out? \\
        \textbf{Reference:} \\
        Q1: When did the new Christopher Robin come out in Burbank? \ A1: July 30, 2018 \\
        Q2: When did the new Christopher Robin come out throughout the United States? \ A2: August 3, 2018 \\
        \textbf{Prediction of \modelname\ w/o Round-Trip Prediction:} (QA1-QA2: \Fanswer=50.0, \Fedit=28.6) \\
        Q1: When did the new christopher robin film come out in the US? \ A1: august 3, 2018 \\
        Q2: When did the new christopher robin film come out at the Disneyland Resort? \ A2: july 17, 2018 \\
        \textbf{Prediction of \modelname:} (QA1-QA3: \Fanswer=80.0, \Fedit=53.6) \\
        Q3: When did the new christopher robin film come out in California? \ A3: july 30 2018 \\
        \textbf{Relevant Snippets of Passages:} (w/ rank from retrieval \& reranking module) \\
        \underline{Rank 1}: "christopher robin" had its premiere in burbank, california on \hl{july 30, 2018}. released in the united states on \hl{august 3, 2018}, by walt disney studios motion pictures, the film grossed over \$197 million. \\
        \underline{Rank 2}: "christopher robin" premiered in burbank, california on \hl{july 30, 2018}, and was released on \hl{august 3, 2018} by walt disney studios motion pictures. \\
        \underline{Rank 18}: for the first time as a disney movie club exclusive on \hl{july 17, 2018} to coincide with its belated 20th anniversary and the live-action "christopher robin" film, released over two weeks later. \\
    \bottomrule
    \end{tabular}
    \caption{
        Predictions generated by \modelname\ from the \dev\ data. 
        We also manually check all the 100 retrieved and reranked passages, and list the answer-relevant passages here.
        However, the listed passages might be different from the passages that annotators search and read during annotation.
    }
    \label{tab:case_full}
\end{figure*}

\section{More Cases from \modelnameshort: Figure \ref{tab:case_full}} \label{sec:more_cases}

\end{document}